# Gesture Recognition with a Focus on Important Actions by Using a Path Searching Method in Weighted Graph

Kazumoto TANAKA

**Dept. of Information and Systems Engineering, Kinki University**
**Higashi-hiroshima, 739-2116, Japan**

**Abstract**
This paper proposes a method of gesture recognition with a focus on important actions for distinguishing similar gestures. The method generates a partial action sequence by using optical flow images, expresses the sequence in the eigenspace, and checks the feature vector sequence by applying an optimum path-searching method of weighted graph to focus the important actions. Also presented are the results of an experiment on the recognition of similar sign language words.

## 1. Introduction

There have been a number of studies on automatic gesture recognition method. One of the main purposes of the gesture recognition is to realize sign language recognition. In order for ordinary persons to communicate with deaf people, the method is needed to translate sign language into natural language.

It is said that there are similar gestures among sign language words, and some part of the gestures (hereafter, they are called "partial actions") play an important role in differentiating those words. For example, in Japanese sign language words, an action of thumb for the word "lend" is very significant for distinguishing the word from the word "return". However, little attention has been given to the gesture recognition that focuses on such partial actions. Thus, this study has developed a recognition method for focusing on important partial actions. The method promises to improve recognition rate of the gestures that have important meanings in their local actions.

Most of studies on gesture recognition have been conducted by utilizing Hidden Markov Models [1][2][3], Dynamic Time Warping [4][5] or Neural Networks [6][7], while the novel recognition method that utilizes a string-matching method based on a path-searching method has been proposed [8]. The target of the recognition method was relatively large motions as a wide hand-swing, not local actions often used in sign language words. However, by modifying the string-matching method, this study has realized a matching method for focusing on important partial actions to recognize sign language words successfully.

In this paper, a generation method of time-series feature vectors of gestures is described first. The feature vectors are employed in the "modified" string-matching. Next, an optimum path-searching method in weighted graph where the cost of the graph edges corresponding to important partial actions is reduced is proposed for realizing our matching method. Also presented are the results of an experiment on the recognition of similar sign language words.

## 2. Gesture recognition by using a path searching method in weighted graph

Since the interest in this study rests on the effects of the matching method by which feature vectors of important partial actions are focused to distinguish similar gestures, we will proceed to discuss single gestures as the study subject. Thus, the author will have the gesture be in a stationary state at the start and end of the action and will not address any special word spotting.

In general, gesture recognition methods utilize time-series feature vectors for pattern matching in a gesture dictionary. The way of feature vector composition can be divided broadly into two categories: one utilizes geometric characteristics in images (model-based method) and another utilizes appearance characteristics of images (appearance-based method) [9][10]. The appearance-based method has merits in representing objects that are difficult to express geometric characteristics. In the study, feature vectors of complex optical flow images from sign language gestures are composed by using the appearance-based method.

As known well, appearance based methods often utilize the eigenspace method for reducing dimensions of the feature vector. The feature vector generation method by using the appearance-based method with the eigenspace is described in 2.1 and 2.2. The method of gesture representation for constructing gesture dictionary and the matching method by modifying the string-matching method are described in 2.3 and 2.4 respectively.

2.1 Generation of partial action sequence

The method generates several partial action images from the optical flow images which are calculated from motion images by using a local correlation operation. This study established the criteria for the generation as the position of flow vectors and the directional change. The generation procedure is given in the following:

Step 1: A raster scan is performed to search already labeled flow vectors on the optical flow image, Image_i (the suffix i is the frame number; the initial value is 1). If such a flow vector is found, labeling by 8-neighbours will be conducted until unlabeled vector is not found within the connected component of the flow. However, if the angle between the labeled vector and its neighbor vector exceeds a given threshold, no labeling is conducted. Once this raster scan is completed, the next raster scan will be performed on the Image_i to search unlabeled flow vectors. If an unlabeled flow vector is found in the scan, new label will be attached and, in a similar manner as the above, labeling by 8-neighbours will be conducted.

Step 2: If the angle between the flow vector at the coordinate (m, n) of Image_i v:=(vx, vy) and the flow vector u at the coordinate (m+vx*dt, n+vy*dt) of Image_i+1 is less than the threshold, the label same as that of v will be attached to u. Here, dt is a sampling time span. This is performed to all flow vectors of Image_i.

Step 3: When i:=i+1 and i is not over the last number of the image series, one will return to Step 1.

Step 4: Each group of flow vectors that have the same label is extracted and then is put in a new frame as a new optical flow image (see Fig. 1). This is performed to all labels at each optical flow image.

Step 5: For all the images created in Step 4, if the number of the images that have the same labeled flow vectors is less than a given threshold, the images will be removed for noise canceling.

Step 6: The images that have the same labeled flow vectors are superimposed on the first image among them in time order at each label. As a result, the images are combined into one image at each label (see Fig. 1). The combined image is called partial action image in the paper.

Step 7: The partial action sequence is generated by arranging the partial action images in label order.

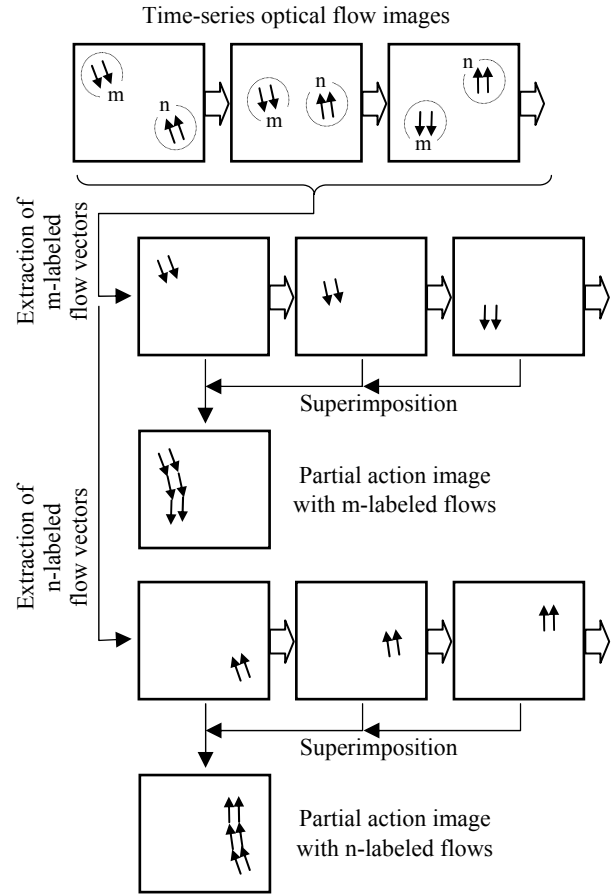

Fig. 1: Generation of partial action image

2.2 Generation of feature vector sequence by the eigenspace method

Because the eigenspace method is sensitive to any position shift of the subject for recognition, it was based on the premise that the gesture would be performed in a nearly identical position inside the window set in the image.

First, the operation described in 2.1 is performed for all dynamic images of each gesture that are prepared for constructing the gesture dictionary. After the operation, for each partial action image, the method generates a vector v by arranging flow vector components in raster scanning order on a partial action image and obtains a set of vector v. The components of the flow vector at each pixel are laid out from component x to component y in the vector.

Next, from all the vector v obtained, the matrix $V := [v_1, v_2, \ldots, v_N]$ is obtained. The eigenspace based on the eigenvectors that correspond to the upper-ranked number k of the eigenvalue is determined, by solving the characteristic equation (1) of the covariance matrix U of the matrix which is obtained by subtracting the mean vector of the column vectors of the matrix V from each column vector.

$$\lambda_i \cdot e_i = U \cdot e_i \tag{1}$$

From this, each vector v of the partial action will be projected by Formula (2) to calculate the feature vector u on the eigenspace.

$$u = [e_1, e_2, \cdots, e_k]^T (v - \bar{v}) \tag{2}$$

Lastly, the feature vector sequence on the eigenspace $<u_1, u_2, \ldots, u_N>$ can be obtained from the partial action sequence.

### 2.3 Gesture representation for constructing gesture dictionary

Each gesture in the gesture dictionary is represented by a cluster sequence on the eigenspace. Each cluster is formed by plural patterns of a feature vector that are obtained from a variation (discrepancy) of a partial action to cope with the problem of the variation. Each cluster is approximated by the k-dimensional normal distribution by Formula (3).

$$f(x;\mu,\Sigma) = \frac{1}{\sqrt{2\pi}^k \sqrt{|\Sigma|}} exp\left[-\frac{1}{2}(x-\mu)^T \Sigma^{-1}(x-\mu)\right] \tag{3}$$

Here, $\mu$ is the mean vector of the cluster, $\Sigma$ is the covariance matrix of the distribution of the cluster. Thus, each gesture in the dictionary will be represented as a sequence of the cluster c represented by the k-dimensional normal distribution as seen in Fig. 2.

$$\{c_1, c_2, \cdots\} \equiv \{\underbrace{f(x;\mu^1,\Sigma^1)}_{\text{Normal distribution of partial action } P_1}, \underbrace{f(x;\mu^2,\Sigma^2)}_{\text{Normal distribution of partial action } P_2}, \cdots \}$$

Fig. 2: Gesture representation in the dictionary

### 2.4 Matching method by a path-searching method in weighted graph

First, Fig. 3 shows the outline of the matching method [8] of one-dimensional feature sequences by using the string-matching method [11] based on a path-searching method. The graph in Fig. 3 have X{A, B, C, A, E, F, G} of the one-dimensional sequence on the horizontal axis and Y{A, H, C, I, F, J} on the vertical axis, and a white circle is attached to the intersection where sequential elements match each other. Added from the intersection is an edge in the lower left diagonal direction. For each edge, a cost is established; the cost is 0 for the added edge in the diagonal direction and the rest are all 1. At this time, obtaining the minimum cost path from the node $V_{00}$ to $V_{76}$, {A, C, F}, which corresponds to the white circle on the edge in the diagonal direction included on that path, is LCS (Longest Common Subsequence) of X and Y, thus the number of elements of LCS $length(LCS(X, Y))$ is 3. From this and Formula (4), the degree of similarity $Sim(X, Y)$ can be obtained. Here, the method uses Dijkstra's method to obtain LCS.

$$Sim(X,Y) := \frac{length(LCS(X,Y))}{max(length(X), length(Y))} \tag{4}$$

Next, the matching method for enabling focusing on important partial actions is described in the followings:

Regarding the sequence of the cluster in the dictionary and the feature vector sequence of recognition subjects, individual elements (corresponding to partial actions) are laid out composing a graph $G_{PQ}$ as in Fig. 3. If the cluster sequence P and the feature vector sequence Q are

$$P := \{c_1, c_2, \cdots, c_P\} \tag{5}$$

$$Q := \{u_1, u_2, \cdots, u_Q\} \tag{6}$$

respectively, which elements will match each other are determined by the thresholding Mahalanobis distance to be defined in Formula (7).

$$d \equiv \sqrt{\left(u_i - \mu^{c_j}\right)^T \Sigma^{c_j^{-1}} \left(u_i - \mu^{c_j}\right)} \tag{7}$$

Here, $\mu^c$ is the mean vector of the cluster c, $\Sigma^c$ is a covariance matrix of the distribution of the cluster. From this matching result, the edge in the diagonal direction is added to $G_{PQ}$, the cost of 1 or 0 is allocated for each edge according to the rules.

The focus on important partial action will be conducted as follows: Let a cluster $c_k$ in the sequence P correspond to an important partial action. If P is laid down on the horizontal axis to $G_{PQ}$, the cost of the edge will be changed by Formula (8).

$$\text{For all } j: j \in \{0,1,\cdots,n\} \quad \text{do} \qquad (8)$$
$$cost\big((V_{k-1\,j}, V_{k\,j})\big) := m + n$$

Here, m and n are the numbers of elements of P and Q respectively, $V_{ij}$ is any node in the graph. From this, if Q has an element that is included in $c_k$, restrictions can be established so that the edge in the diagonal direction toward that matching node can be selected as a path. In the example in Fig. 3, if the fourth element of Sequence X, Element A, corresponds to an important partial action, LCS is {A, F} and *length*(*LCS*(X, Y)) will be 2.

From the graph composed in the above, one can obtain *LCS*(P, Q) using the Dijkstra method and search in the dictionary based on *Sim*(P, Q). In the study, programming was done by the creator of the dictionary designating the important partial actions and attaching a flag to that data.

each word and the optical flow images were obtained to create a gesture dictionary for basic 16 words including the four words. The eigenspace was decided as 4-dimensional space to represent each feature vector of the words.

The equipment used for image processing was a personal computer (CPU: Intel(R) Core(TM)2 Duo CPU, 2.20GHz; system memory 3.07GB; OS: Windows XP). The library of OpenCV was used for image processing. The resolution for image processing was 320 x 240. The entire upper body of the subject was imaged in front of a blackout curtain. The video rate was 60fps.

Fig. 4 shows an example of the optical flow image when performing a gesture of "lend". An extracted optical flow image of the thumb and that of the hand without the thumb are also shown in the figure. The partial action of the thumb is an important feature to distinguish "lend" from "return (home)." An example of the partial action image of the thumb is shown in Fig. 5. The partial action image was generated by combining the same labeled optical flow images of the thumb.

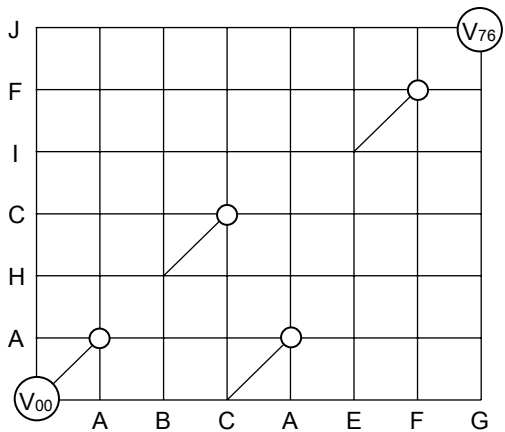

Fig. 3: A Graph for the matching method of 1D feature sequences

## 3. Experiment

To verify the efficacy of this method, the study conducted a recognition test on Japanese sign language words. The words used in this experiment included "say," "order," "return (home)," "lend." Of them, "say and order" and "return (home) and lend" are very similar in action respectively. Two subjects made the action 30 times for

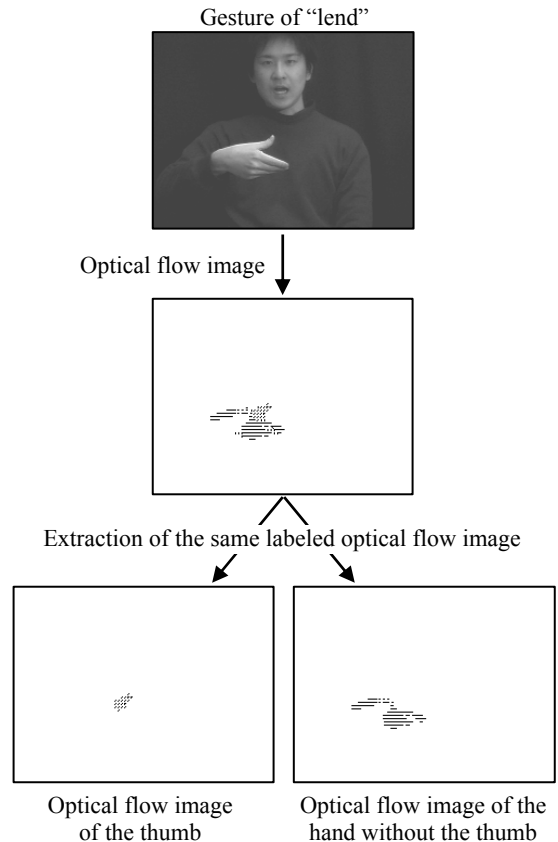

Fig. 4: Optical flow images when performing a sign language word

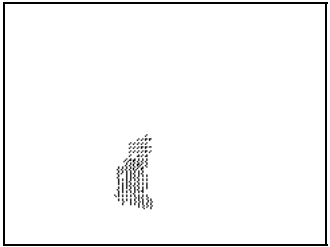

Fig. 5: Partial action image of the thumb

For the recognition experiment, a total number of four people, including two subjects at the time of creating the gesture dictionary, performed a gesture 20 times per word. Table 1 shows recognition rates of the four words and the average recognition rate of the basic 16 words actions by using the proposed method. Table 2 shows recognition rates when important features were not focused. A and B in the tables denote the subject group and the other group, respectively, at the time of creating the dictionary.

Table 1: Recognition rates with the proposed method

|  | A | B |
|---|---|---|
| "say" | 69.0% | 63.5% |
| "order" | 70.5% | 65.0% |
| "return" | 77.0% | 70.5% |
| "lend" | 76.0% | 67.5% |
| Average | 79.5% | 68.5% |

Table 2: Recognition rates without important feature focusing

|  | A | B |
|---|---|---|
| "say" | 58.0% | 51.0% |
| "order" | 60.5% | 49.0% |
| "return" | 62.5% | 52.5% |
| "lend" | 64.5% | 57.0% |
| Average | 76.5% | 65.5% |

## 4. Discussion

The results of the experiment show that the recognition rate with focused important features was greater than that without focusing. Thus, the efficacy of this method was confirmed. Furthermore, the following points were clarified:

The recognition rate of "say" and "order" was lower than that of other words. The difference between the two words was the difference in the action near the starting location. However, since the variance of the starting location was great, the matching of this important partial action often failed. Thus, it is needed to improve the robustness against the location shift for the future work.

The reason for the recognition rate of Group B being lower than that of Group A was the difference in individual "habits." For this reason, it will be necessary to obtain gesture patterns from a large number of people and improve the model accuracy of the partial action cluster. For the model, the distribution expression of "habits" by Gaussian Mixture Model can be used.

Overall recognition rates are lower than those in the previous studies of sign language word recognition. This is because the study did not use shapes of the hands and fingers, which are important pieces of information. However, the purpose of this study was to make it possible to recognize gestures that have important local actions. Therefore, in this respect, the author believes that the purpose has been achieved.

## 5. Conclusion

For the identification of similar gestures, the paper proposed the gesture recognition method with a focus on important local actions. The method generates the partial action sequence by using optical flow images, expresses the sequence in the eigenspace, and checks the feature vector sequence by applying the optimum path-searching method of weighted graph. The paper also showed the efficacy of this method by conducting a recognition test on similar sign language words. For the future, in addition to the improvement on the robustness against the location shift, the author plan to tackle the issue of employing this method for sign language and gestures that include body parts other than hands (shaking of the head, etc). Future work will involve further verification of the method with various sign language words.

### Acknowledgments

This work was partially supported by a Grant-in-Aid for Scientific Research (c) 20500856 from the Japan Society for the Promotion of Science.

**Kazumoto T.** is an associate professor of Faculty of Engineering, Kinki University, Japan. He received B.S. degree from Chiba University in 1981, and Ph.D. degree from Tokushima University in 2002. In 1981 he joined Mazda Motor Corporation, he was engaged in research and development of robot vision technology. His current interest is mainly gaming technology. He is a member of Japanese Society for Information and Systems in Education and Institute of Electronics, Information and Communication Engineers.